\documentclass[10pt,letterpaper]{article}
\usepackage[top=0.85in,left=2.75in,footskip=0.75in]{geometry}

\usepackage{amsmath,amssymb}

\usepackage{changepage}

\usepackage{textcomp,marvosym}

\usepackage{cite}

\usepackage{nameref,hyperref}

\usepackage[right]{lineno}

\usepackage[nopatch=eqnum]{microtype}
\DisableLigatures[f]{encoding = *, family = * }

\usepackage[table]{xcolor}

\usepackage{array}

\newcolumntype{+}{!{\vrule width 2pt}}

\newlength\savedwidth

\raggedright
\usepackage[aboveskip=1pt,labelfont=bf,labelsep=period,justification=raggedright,singlelinecheck=off]{caption}

\makeatletter
\renewcommand{\@biblabel}[1]{\quad#1.}
\makeatother

\usepackage{lastpage,fancyhdr,graphicx}
\usepackage{epstopdf}
\fancyheadoffset[L]{2.25in}
\fancyfootoffset[L]{2.25in}
\begin{document}
\vspace*{0.2in}

\begin{flushleft}
{\Large
\textbf\newline{Vision Transformer attention alignment with human visual perception in aesthetic object evaluation} 
}
\newline
\\
Miguel Carrasco\textsuperscript{1*\ddag},
César González-Martín\textsuperscript{2\ddag},
José Aranda\textsuperscript{3\ddag},
Luis Oliveros\textsuperscript{3\ddag},
\\
\bigskip
\textbf{1} Escuela de Informática y Telecomunicaciones, Universidad Diego Portáles, Santiago, Chile
\\
\textbf{2} Department of Specific Didactics, University of Cordoba, Cordoba, Spain
\\
\textbf{3} Facultad de Ingeniería y Ciencias, Universidad Adolfo Ibáñez, Santiago, Chile
\\
\bigskip

%
%

\ddag These authors also contributed equally to this work.



* miguel.carrasco@mail.udp.cl

\end{flushleft}
\section*{Abstract}
Visual attention mechanisms play a crucial role in human perception and aesthetic evaluation. Recent advances in Vision Transformers (ViTs) have demonstrated remarkable capabilities in computer vision tasks, yet their alignment with human visual attention patterns remains underexplored, particularly in aesthetic contexts. This study investigates the correlation between human visual attention and ViT attention mechanisms when evaluating handcrafted objects. We conducted an eye-tracking experiment with 30 participants (9 female, 21 male, mean age 24.6 years) who viewed 20 artisanal objects comprising basketry bags and ginger jars. Using a Pupil Labs eye-tracker, we recorded gaze patterns and generated heat maps representing human visual attention. Simultaneously, we analyzed the same objects using a pre-trained ViT model with DINO (Self-DIstillation with NO Labels), extracting attention maps from each of the 12 attention heads. We compared human and ViT attention distributions using Kullback-Leibler divergence across varying Gaussian parameters ($\sigma=0.1 to 3.0$). Statistical analysis revealed optimal correlation at $\sigma=2.4\pm0.03$, with attention head \#12 showing the strongest alignment with human visual patterns. Significant differences were found between attention heads, with heads \#7 and \#9 demonstrating the greatest divergence from human attention ($p\leq 0.05$, Tukey HSD test). Results indicate that while ViTs exhibit more global attention patterns compared to human focal attention, certain attention heads can approximate human visual behavior, particularly for specific object features like buckles in basketry items. These findings suggest potential applications of ViT attention mechanisms in product design and aesthetic evaluation, while highlighting fundamental differences in attention strategies between human perception and current AI models.


\section*{Introduction}
Human visual attention is a crucial process that allows individuals to focus on specific visual stimuli, filtering information from the environment, necessary due to the biological limitations of processing all the visual inputs we receive \cite{carrasco2011}, which is essential for human perception \cite{lai2021} and affects their behavior \cite{capozzi2024,theeuwes2010}.
Prior to attention, preattention occurs, a selective attention where some inputs are weighted over others, and the weights must be chosen for specific objectives \cite{reeves2020}. For this, an analysis of visual characteristics \cite{wilkinson2024} or low-level features (e.g., color, shape, orientation) \cite{wang2019}, and their location in space \cite{bouchard2025}, also called mid-level features \cite{gordo2015,mikhailova2022}, takes place. Without instruction to the observer, the contrast between the visual characteristics of an object and the other components of the scene appears to be determinant in guiding attention \cite{wolfe2019}. In short, singularities in images are the characteristics that determine visual attraction \cite{obeso2022}.
Visual attention is composed of two types of mechanisms: overt, moving the eyes toward a specific object, and covert, which is when attention is focused on a peripheral zone, voluntarily or involuntarily without directing the gaze there, with the latter preceding the former \cite{meur2020,posner1980}. In turn, visual attention is categorized into two functions: Bottom-up, which is initial attention produced by salient stimuli in the environment, and top-down attention, which is captured by the relevance, objectives, intentions, context, and prior knowledge of the observer \cite{katsuki2014,lai2021,rensink2000,corbetta2002}. However, this dichotomy is debated \cite{awh2012} as they are two neurocorrelated processes \cite{katsuki2014}.

During visual attention processing, effort is required to maintain focus on the stimulus, whether intentionally or automatically \cite{scholl2001}. Contrary to common assumptions, visual attention operates as a slow, rhythmic process \cite{ward1996} that varies depending on the object's location within the representational space \cite{jia2017}. This variability is evident in findings showing that the initial fixation does not determine the subsequent course of action \cite{zhang2025}. The importance of observation time on an object has been studied as a strong predictor of purchase \cite{behe2015,clement2007,li2024}, and it is deduced that the more you like an object, the longer you look at it, which increases the possibility of purchase \cite{gidlof2017}. However, while some theoretical currents maintain that the purchase decision occurs after fixations, others point out that this action takes place during fixations \cite{orquin2013}.
In this sense, aesthetics plays an important role in visual exploration \cite{grinde2022} through visual characteristics, such as orientation, luminance, size, color, or shapes, positively influencing the speed of visual search \cite{guo2020}, capturing and preserving visual attention more effectively \cite{rolke2019} and therefore, fixations, which improves perception and is related to emotions \cite{pearce2016,vartanian2004}, regardless of conditions \cite{fuchs2011} and the nature of the object \cite{goller2019,meur2020}. The correlation between visual attention and aesthetic preference has been studied through faces \cite{mitschke2017}, objects \cite{farzanfar2023}, architecture \cite{zhang2018}, or works of art \cite{straffon2022}, demonstrating that it affects self-relevance \cite{macrae2018}. However, to the extent of our knowledge, visual attention in artisanal production has not been explored in depth, where the aesthetic dimension is also a determining factor for its consumption. One could cite the work of S. Zhang \cite{zhang2022}, who shows the aesthetic influence of plates on food

Regarding visual attention analysis techniques, eye tracking technologies have predominated in recent years, achieving extensive research development across different fields \cite{blascheck2017,carter2020}, including artistic objects \cite{koide2015,marin2022} of various typologies and styles \cite{gartus2015,leder2004,meur2020,mitschke2017,zangemeister1995}, being an ideal tool for studying visual attention \cite{carter2020}. However, the recent emergence of the Deep Learning model called Vision Transformers (ViT) has revolutionized the field of computer vision and automatic image processing, equaling or surpassing other computational models such as Convolutional Neural Networks (CNNs) by using image patches and attributing positional embeddings to them, passing through the encoder independently, which allows it not to lose information about their order \cite{khan2022}. Within the encoder, these patches pass through an attention module that contains multi-head attention layers that achieve the so-called self-attention characteristic of ViT. Its peculiar structure gives it a variety of unique characteristics, highlighting the ability to incorporate global and local information in the lower layers of the network \cite{raghu2021}. Additionally, they manage to create shortcuts between their neurons that facilitate connections and performance, allowing it to have an understanding of the complete context of the image and from the beginning can classify even when the image pieces are not delivered in the correct order, unlike CNN models, which depend on initial layers focused only on local information. This self-attention mechanism was originally proposed in the Transformer model by Vaswani et al. \cite{vaswani2017}, allowing simultaneous relation of all input regions at different levels of spatial hierarchy.
However, it cannot be well explained how ViT determines the attention of each part of the images it classifies. Understanding this depends on a large number of neurons in the model and a black box effect occurs where it is impossible to see the steps taken to reach the result. Recent studies have observed that, although Vision Transformers manage to capture perceptual groupings similar to humans, they tend to assign relevance differently, sometimes highlighting distractors or secondary elements \cite{tuli2021}.

Studies such as Raghu et al. (2021) \cite{raghu2021} have delved into understanding the functioning of ViTs compared to CNN models. Given the difficulty, the study was based on Central Kernel Alignment (CKA), which provides a scalar value that can be used to determine quantitative similarity between different layers more easily. When applying CKA to ViT and ResNet, it was determined that their first 60 layers were similar, but later in the upper layers they differed considerably. Additionally, ViT layers change uniformly while ResNet layers had an abrupt change between lower and upper layers. The functioning of multi-head attention layers was also analyzed by restricting the distances they covered. In this way, it was discovered that they provide global information even in the lower layers, which differed completely from CNNs where the first layers contain only local information. Even when implementing tokens to ResNet that represent convolutional channels of a particular spatial zone to compare their functioning with ViT attention, it was observed that these focus better on the image and its contour compared to CNNs that use more of the image's texture for classification. Despite the above, they were able to discover several characteristics of how self-attention functions. They noted that their methodology based on the use of CKA could be deepened with finer methods. In other studies such as Tuli et al. (2021) \cite{tuli2021}, they have delved into the problem using other metrics. In this case, precision and error when classifying the same set of images, they found that ViTs are more similar to humans than CNNs. Even so, a new perspective on the internal functioning of ViT attention could not be provided. On the other hand, there is the possibility of deepening knowledge of multi-head attention layers through analysis of their attention and working inversely, from the result in images toward the internal structure of these. Due to this, it is appreciated that the flexibility of self-attention in ViT is closer to human vision. In this aspect, the study agrees that ViT better explains human visual attention during reading than the computational E-Z Reader model.

Conversely, there are studies that demonstrate gaps between traditional/CNN-based saliency models, deep neural networks, and human performance in visual processing, showing that ViTs tend more toward perceptual grouping than attention, which approximates the behavior of lateral interactions in the human visual cortex \cite{kotseruba2024}. On the other hand, Mehrani and Tsotsos (2023) \cite{mehrani2023} demonstrate that ViTs assign relevance to elements differently from human attention, highlighting distractors or elements located in the background in the results. Along this line, they point out that human visual attention involves both feed-forward and feedback mechanisms, while in ViTs feed-forward mechanisms predominate, suggesting fundamental differences in how attention is implemented. It has been proposed that a key difference between human attention and that of ViTs lies in the combination of feedforward and feedback mechanisms in humans, while in ViTs a primarily feedforward architecture predominates, limiting their approximation to natural visual processing, performing more global attention \cite{caron2021}.

Given the discrepancies regarding the similarity in attention performed through ViT and human attention, this research proposes to study the level of correlation in attention when viewing the same sample composed of two typologies of artisanal objects (bags made through basketry and ginger jars). The choice of these productions responds to the continuation of using artistic objects for attention studies due to their visual characteristics and aesthetic components. Furthermore, the morphology of the two artisanal productions is opposite, as while bags tend toward rectilinear forms and polyhedral structures, ginger jars tend toward the vertical and their forms are curvilinear. This will allow us to make a comparison and detect variability in the results regarding attention.
The study of similarity between the two attentions (human vs ViT) will allow for deeper exploration of the use of this technology in the creation and design process of commercial products, for detecting visual attraction zones, thus allowing knowledge of the level of visual attraction in advance. Therefore, this research establishes the following hypotheses:

\begin{itemize}
\item H1. The Vision Transformer attention module and human visual attention do not present statistically significant differences
\item H2. ViT is an applicable technology in artisanal product design for detecting aesthetic interest zones.
\end{itemize}

To respond to the proposed hypotheses, the following objectives are established:
\begin{itemize}
\item O1. Statistically determine the correlation between ViT and human attention in a dataset of images of artisanal products
\item O2. Analyze visual interest zones in artisanal objects with both attention mechanisms (ViT and human)
\end{itemize}

\section*{Materials and methods}
\begin{figure}[!h]
\centering
    \includegraphics[width=1\linewidth]{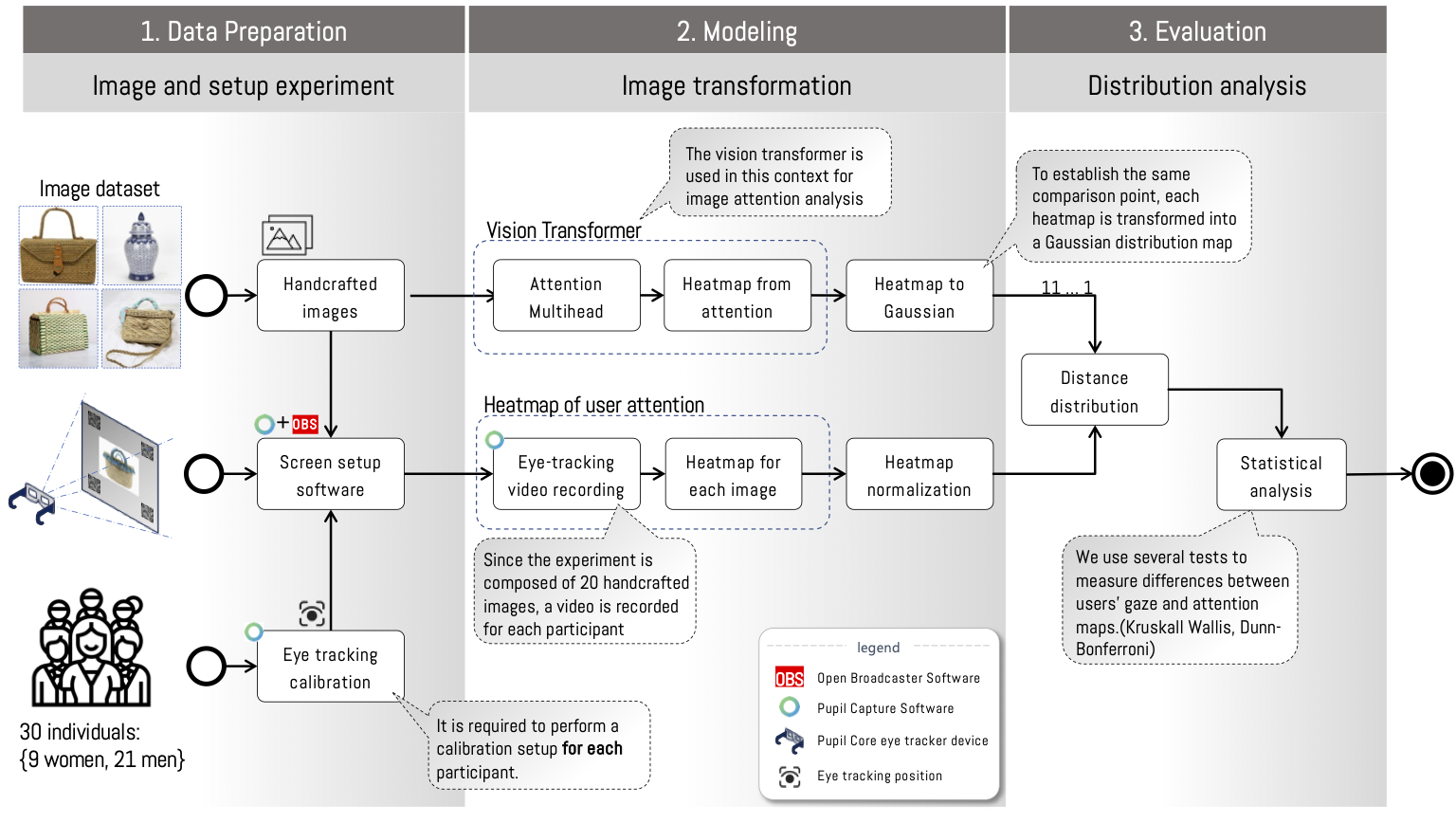}
\caption{{\bf General diagram of the experimental attention analysis process}. The process is composed of three stages: 1) data preparation and experimental setup, 2) image transformation, and 3) distribution analysis. Data preparation and setup consists of the image evaluation process by each experiment participant. This stage requires the use of an eye-tracker to determine the gaze position of experiment participants and define the experimental conditions. Image transformation consists of generating an attention map through the ViT attention module and experimentally by users on a set of objects. The last stage performs the comparison between both information sources and thus determines if there is any type of correlation}
\label{fig1}
\end{figure}

The methodology is composed of three stages defined as data preparation, modeling, and evaluation. Each of them is composed of sub-components that allow the integration of the experiment through the flow of information between software and experimental components (see Fig.~\ref{fig1}). Below we explain each of the stages in detail.

\subsection*{Data Preparation}
\begin{figure}[!h]
\centering
    \includegraphics[width=1\linewidth]{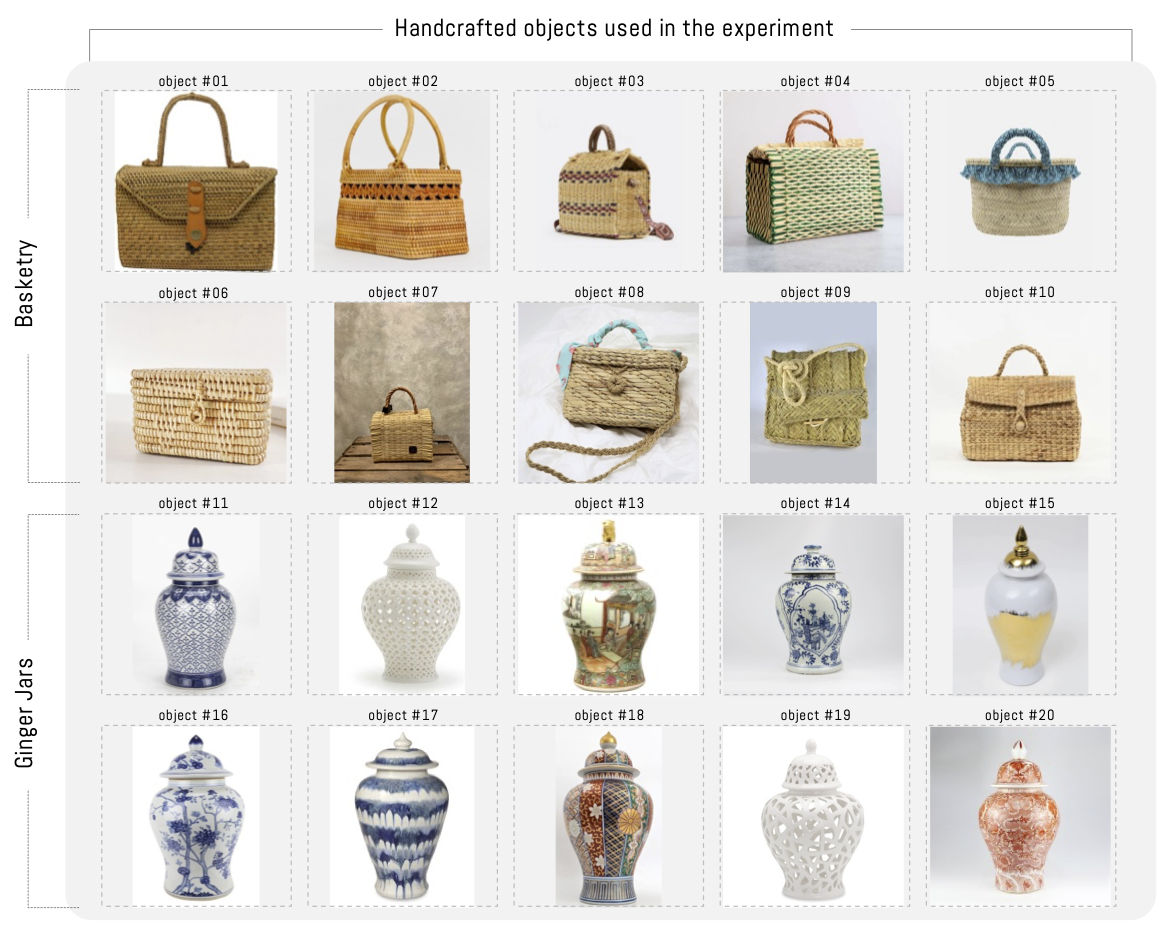}
\caption{Objects used in the experiment composed of ten basketry objects and ten ginger jars. The objects were randomly selected taking into consideration that there should be the fewest possible objects in the background.}
\label{fig2}
\end{figure}

The experiment consists of viewing a group of images by a set of people in a controlled environment through an eye-tracker. The analyzed objects correspond to craft pieces, specifically basketry and ginger jars which form part of the RRRemaker project\footnote{RRREmaker is a project funded by the European Union focused on the reuse of craft objects through generative intelligence techniques.}. The selection was composed of ten basketry objects and ten jars, which can be seen clearly and without relevant external visual distractors (see Fig.~\ref{fig2}). The selected objects vary slightly in sizes and decorations, maintaining unity in their materials, colors and shapes, to reduce distractions and force attention toward details. In the case of the jars, these have curvilinear forms and common structures, but with variations mainly highlighted in decorations (colors, figures and shapes). In both cases, scene distractors have been reduced and all are free of logos and text.

\begin{figure}[!h]
\centering
    \includegraphics[width=1\linewidth]{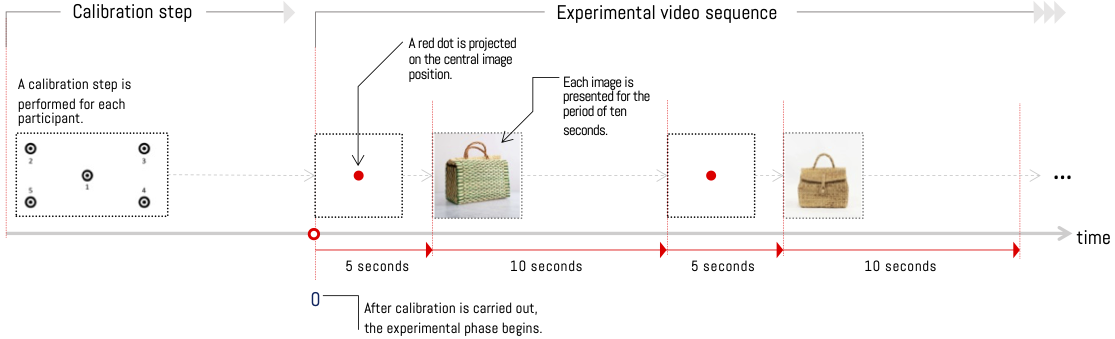}
\caption{{\bf Experimental procedure for object visualization}. Before starting the experimental phase, a calibration procedure is performed with the recording of a sequence of five points on the screen. Once this process is completed, the experimental phase begins through the projection of an image with a white background and red dot which is displayed for 5 seconds. Then one of the 20 objects is displayed for 10 seconds. This procedure repeats until all objects have been displayed.}
\label{fig3}
\end{figure}

To record the visual information from participants, we have used a Pupil Core model eye tracker from Pupil Labs manufacturer through the Pupil Capture software. This software allows recording from multiple device sensors, such as microphone, front camera, and pupil refraction cameras. Additionally, the software performs camera synchronization for eye tracking calculation through a calibration process (Fig.~\ref{fig3} calibration step). This process allows precise determination of the user's gaze on the experiment screen. To do this, it relates the gaze position on the screen with the eye position relative to the eye-tracker's internal camera.

The image sequences displayed have been carried out through the following procedure. First, a red dot is presented at the center of the screen for five seconds. Then each of the 20 objects is presented for ten seconds. Then it returns to the first step to transition object by object from the basketry set and then jars (Fig.\ref{fig3}). The objective of looking at a red dot during the transition between objects seeks to center the gaze in the same position at the beginning of displaying a new object. In this way, we reduce the error of gaze position by being situated in another position during the transition. On the other hand, displaying the image for 10 seconds was chosen as it is a time range that allows human attention to understand the most relevant information, decreases fixation generated by mental load, and provides a good amount of effective fixation samples\cite{locher2007visual}.

\begin{figure}[!h]
\centering
    \includegraphics[width=1\linewidth]{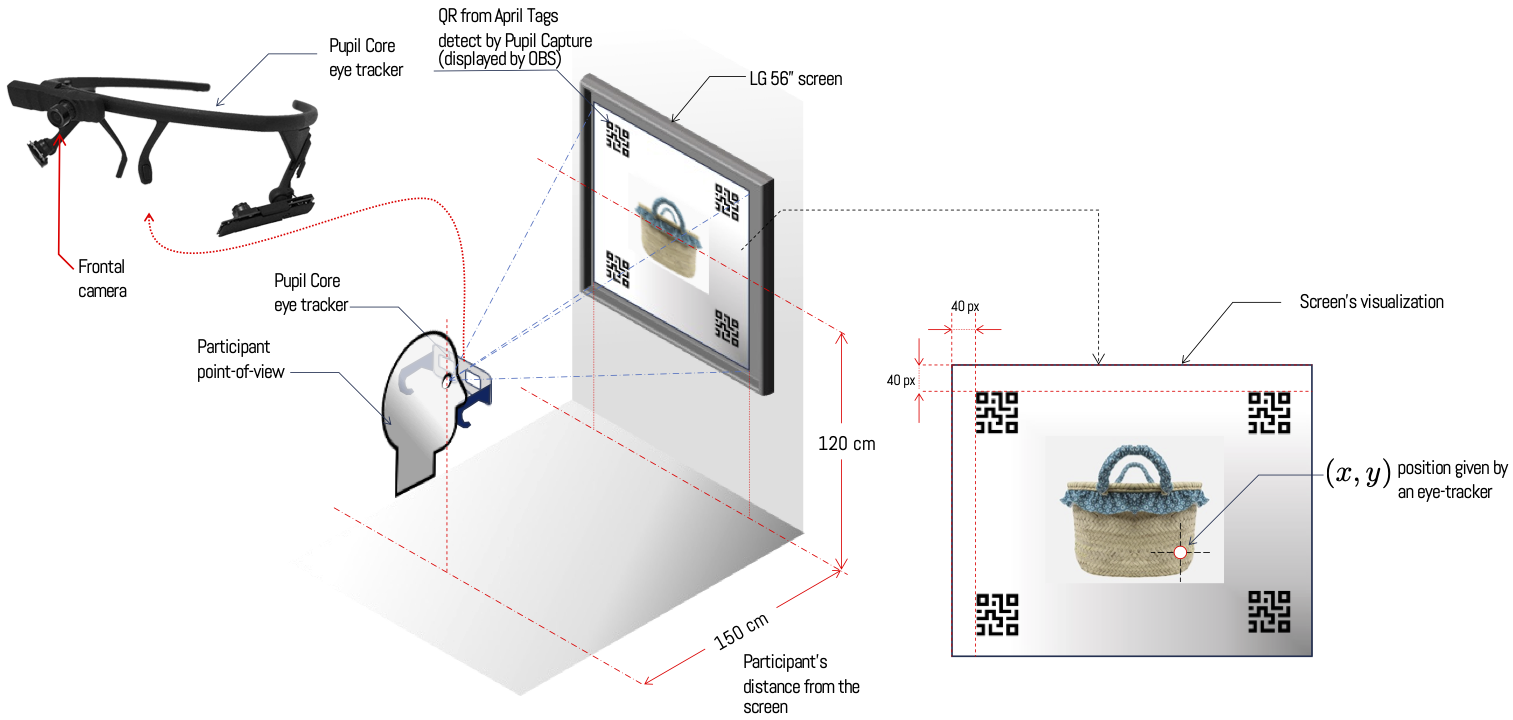}
\caption{{\bf Participant setup in front of the screen during the experimental phase}. All participants remain seated while the experiment is conducted. At the beginning of each experiment, a calibration process is performed with the eye-tracker, along with explaining the experiment to the participant. The chosen distance between the user and screen remains relatively fixed at 150 cm as it reduces visual fatigue.}
\label{fig4}
\end{figure}

The experiment uses a reference system that allows real-time location of gaze position on the screen. For this, it is necessary to employ four QR markers of the April Tags type which are recognized in real time by the Pupil Capture software and allow precise determination of gaze position on a reference system. This allows participants to make movements with their body and head with complete freedom, and the software to detect gaze position relative to the screen. This procedure has been carried out in combination with OBS (Open Broadcaster Software) which displays a QR code in each corner of the screen, along with object display through an application developed in Python. On the other hand, the distance between the screen and user has been maintained at 150 cm as it allows obtaining a large part of the participant's visual field, minimizing visual fatigue (see setup in Fig.~\ref{fig4}). This procedure has been implemented on an Asus TUF 15 computer with 16gb of RAM, Intel Core i5 CPU and an NVIDIA 1650 graphics card, and images have been projected on an LG screen with 56-inch resolution and 60 Hz refresh rate.

In order to obtain statistically valid results, the project has collected information from 30 participants, who voluntarily accepted the experiment and signed informed consent. The selected participants meet the following criteria: 1) being persons over eighteen years of age, 2) not reporting any pathology or ocular deficiency that would prevent them from viewing images at distances less than 150 cm away, and 3) not reporting any type of pupil refraction that would make it unfeasible to use an eye-tracker device with this type of technology. An exception was made if the condition can be corrected through contact lenses, as these do not alter the measurements of the eye tracker used. Additional details were recorded for all participants, such as age, gender, and area of knowledge or profession in order to analyze if there is any relationship between these factors and the visual attention of each examined individual.

\subsection*{Modeling}
\begin{figure}[!h]
\centering
    \includegraphics[width=1\linewidth]{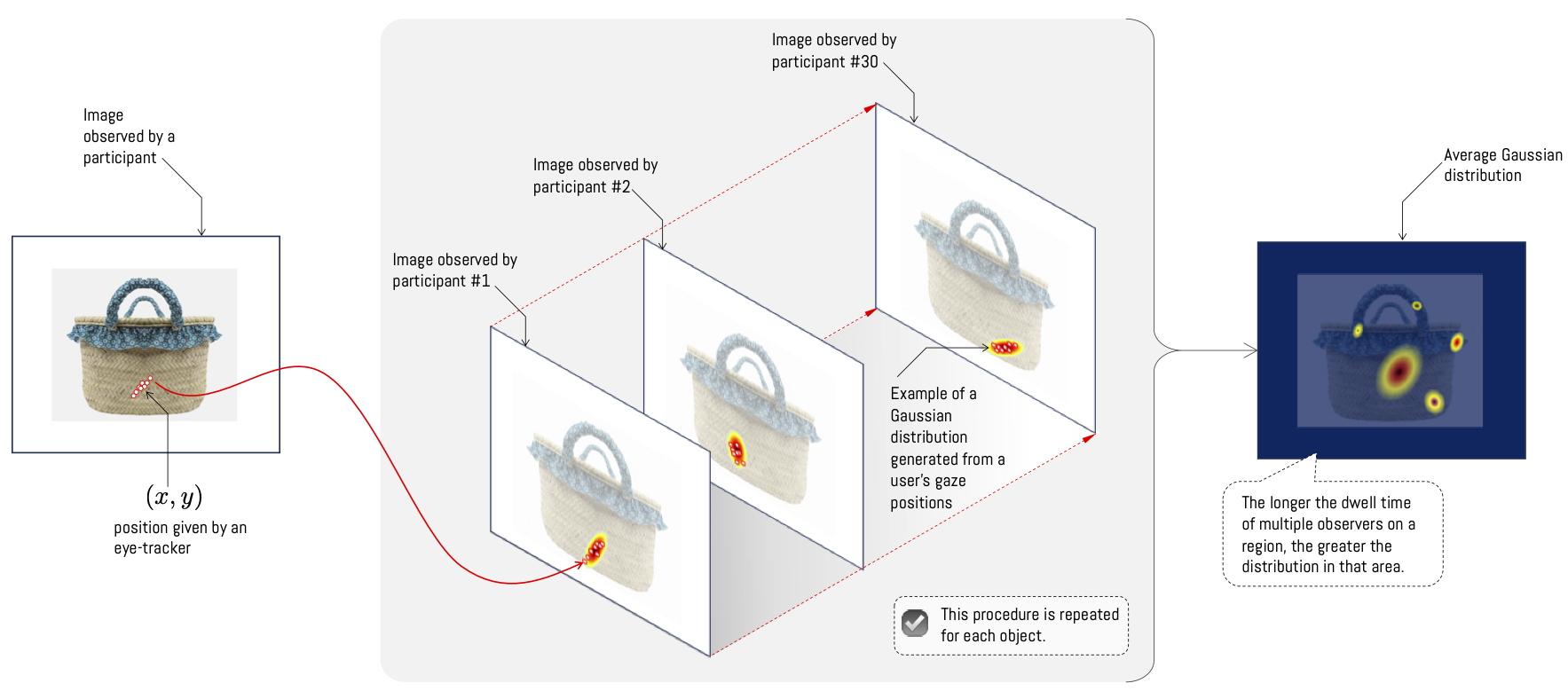}
\caption{Heat map generation according to positions recorded by each observer. The heat map of each object is constructed as the average of individual visualizations transformed to a two-dimensional Gaussian distribution}
\label{fig5}
\end{figure}

"The data generated by the procedure described in Fig.\ref{fig3} are processed using Pupil Player Software, allowing obtaining the gaze position of each participant; which are expressed as coordinates $(x,y)$ in relation to the viewing area (Fig.
\ref{fig5}). Additionally, it is possible to obtain the timestamp, the corresponding frame of the recording, the pupil position coordinates, and the confidence level of the measurement. Thanks to the time and gaze position information, it is possible to estimate a Gaussian distribution over each coordinate taking into consideration the relative time of permanence at a determined location, which allows generating a heat map. The longer the gaze remains on a certain position in the image, the density in that region increases (see example in Fig.\ref{fig5}). This process is performed by each participant on the set of images, obtaining as a result a set of Gaussian distributions in each of the images; each distribution independent of the result of other observations. In this way, the distributions are considered independent, and it is possible to consider their average as the result of observation for each image (Fig.\ref{fig5}., Average Gaussian distribution example). Finally, in order to compare this result with the Vision Transformer attention module, the resulting distribution for each experiment image is normalized.

For Vision Transformer (ViT) training, the pre-training available in Facebook Research's DINO (Self-DIstillation with NO Labels) repository has been used. This research uses a training scheme that consists of training a ViT using another in a teacher-student relationship. Specifically, a process known as self-distillation is applied in which two ViTs pre-trained with the same dataset without categories (ImageNet) but with different parameters are used. The same images are passed through each one with the difference that for the student they are segments of at most 50\% of the image while the teacher has an equal or greater percentage. Then, the teacher ViT weights are updated using those of the student through the Exponential Moving Average (EMA) technique, and then centering is applied to the teacher with a mean over the batch to avoid dominance of any found feature. Finally, the softmax function is applied to calculate the cross-entropy between both networks so that both maintain the same distribution. This form of training achieves better performance and classification power than supervised training, in addition to superior segmentation of relevant zones which allows for clearer visual contrast with human attention~\cite{mathilde2021}.

The architecture used generates as a result 12 independent distributions, each stored in one head of the attention module. In this way, it is possible to consider the average result of the 12 heads (similar to the analysis applied to participants), or consider each of the distributions as an independent result. To visualize the attentions generated by the ViT, specific functions were added that extract the weights obtained by each head of the attention module. In the same way as the attention transformation process was performed on participants, we have transformed each of the weights from the 12 heads to Gaussian distributions so that it is possible to analyze the differences between both.

\subsection*{Evaluation}
To analyze the differences between the heat maps generated by ViT attention and the average visualization of participants, we evaluate the distance between distributions. This comparison is performed through the Kullback-Leibler distance~\cite{kullback1997information}. The Kullback-Leibler distance, also called relative entropy, is a distance between distributions that is relative to one of them, in other words one is used as a reference point. A simple percentage difference calculation can tend to be more biased toward distributions with larger values and, therefore, Kullback-Leibler applies the logarithmic function to eliminate this bias. It is then multiplied by the distribution used as a reference point to apply the difference to each variable of the distribution.

\begin{eqnarray}
\label{eq:KL}
	DKL(P(x)||Q(x)) = \sum_{x \in X} P(x) \log\left(\frac{P(x)}{Q(x)}\right)
\end{eqnarray}

From Eq.~(\ref{eq:KL}) we determine the average distance of participants' attention (see Fig.5: average Gaussian distribution) with each head of the attention module. The objective consists of determining if there is any head where the distance between the user's gaze is similar to one of the heads of the attention module. For this, we use a statistical test that measures the difference between medians, which requires that the samples be independent, continuous, and of the same size. To determine which heads distribute differently, the Tukey Honestly Significant Difference (HSD) test was applied as a post-hoc test to each combination of possible head pairs with the same significance level.

To identify atypical heads, we have determined the p-values obtained to find the heads that achieved the lowest values in each combination and whether they did not exceed the null hypothesis through the HSD statistical test. In this way, it is possible to determine if the null hypothesis is refuted and, therefore, there is at least one different distribution.

\section*{Results}
This section presents the results obtained from the previously outlined methodology. We separate the analysis into participant visualization, ViT model results, and subsequently discuss the fundamental problem of this research, focused on the analysis of similarity and/or difference between ViTs and the average perception of participants.

\subsection*{Sociodemographic information of participants}
The experiment was conducted in the Neuroscience Laboratory of the School of Psychology at Universidad Adolfo Ibáñez (UAI) in Chile and has the approval of this institution's ethics committee (certificate 57/2023). In total, 30 people participated who signed informed consent for conducting the experiment. Each of the participants agreed to provide sociodemographic information such as age, gender, and area of study knowledge (see Table~\ref{table1}). Of the total of 30 participants, 30\% correspond to the female gender and 70\% to the male gender, with an average age of 25.2 years $(SD=4.87)$ for the female gender and 24.3 years (SD=3.62) for the male gender, with a total average of 24.6 years $(SD=3.97)$. Regarding the area of knowledge they ascribe to, 53.3\% is associated with the engineering and sciences area (mathematics, data science, computer science), 36.7\% with social sciences and arts (law, humanities, arts), and 10\% with the business area (marketing, business administration).

\begin{table}[!ht]
\centering
\caption{Sociodemographic information of participants}
\begin{tabular}{l|rrrr|rrrl|rrrr}
       & \multicolumn{4}{c|}{\bf Count} &   \multicolumn{4}{c|}{\bf Mean (years)} &   \multicolumn{3}{c}{\bf $\sigma$} \\
       \hline
 &
  BSN &
  SE &
  SSA &
  $\Sigma$ &
  BSN &
  SE &
  SSA &
  $\mu$ &
  BSN &
  SE &
  SSA &
  All\\ \hline
Female & -    & 6    & 3    & 9      & -   & 26,5 & 22,7 & 25,2 &   -      & 5,6   & 0,6 & 4,87 \\ 
Male   & 3    & 10   & 8    & 21     & 27  & 24,4 & 23,1 & 24,3 &   7,8    & 1,1   & 3,7 & 3,62 \\ \hline
Totals & 3    & 16   & 11   & 30     & 27  & 25,2 & 23,0 & 24,6 &   7,8    & 3,5   & 3,1 & 3,97\\ \hline
\end{tabular}
\begin{flushleft} Note: {\bf BSN}: Business, {\bf SE}: Sciences and Engineering, {\bf SSA}: Social Sciences and Arts
\end{flushleft}
\label{table1}
\end{table}

\subsection*{Participant visualization}
\begin{figure}[!h]
\centering
    \includegraphics[width=1\linewidth]{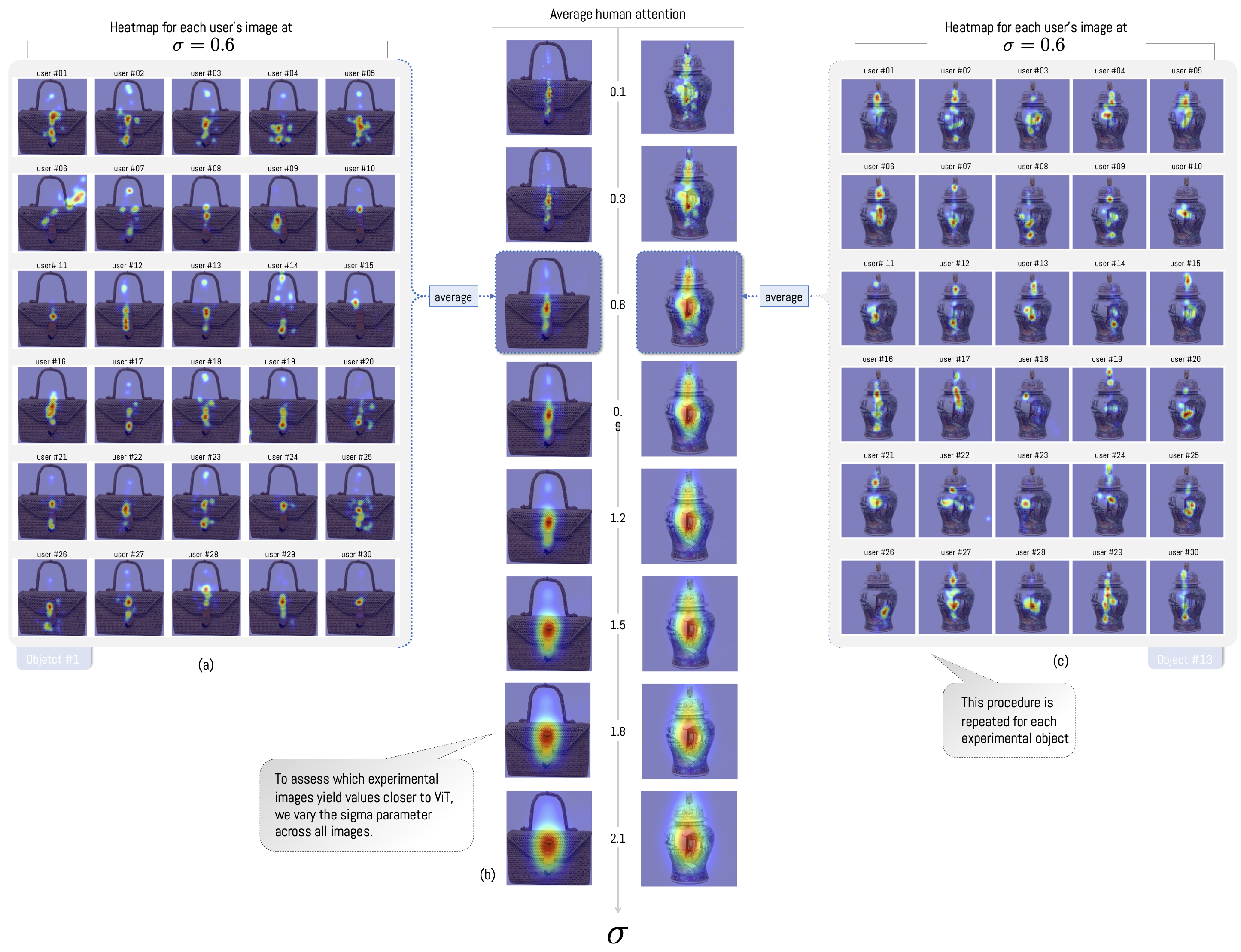}
\caption{(a) heatmap of each user for object \#1 (basketry), (b) average of participants' gaze on an experiment image. As the parameter $\sigma$ increases, the greater the coverage area of the average vision. (c) heatmap of each user for object \#13 (ginger jar).}
\label{fig6}
\end{figure}

The positions recorded by the eye-tracker allow constructing a heat map that identifies the regions where each user has maintained their gaze on the screen. Thus, 600 heatmaps were obtained (30 participants by 20 observed objects). Although Pupil Capture software generates as a result a position in terms of coordinates $(x,y)$, the density can be subsequently modified through the estimation of a two-dimensional Gaussian distribution as a function of the points visualized on the screen (see Fig.\ref{fig5} and Fig.\ref{fig6}). This modification is performed by modifying the parameter $\sigma$, which defines the standard deviation of the Gaussian distribution (see change of $\sigma$ in Fig.\ref{fig6}b). In this way, as the parameter increases, so does the area and density of the zones where the average gaze of the study participants is centered. On the contrary, when $\sigma$ is low, we obtain isolated regions with low density (see example Fig.\ref{fig6} when $\sigma=0.1$). To obtain the average heat map, it is necessary to sum the distributions generated by the 30 users (for the same object), and then normalize this result making the sum equal to 1.0. This process is repeated each time the parameter $\sigma$ varies, totaling 30 variants of $\sigma$for each object, where $\sigma\in\{0.1,0.2,0.3,...,3.0\} = \{0.1\times k | \space k \in \mathbb{N},\space 1\leq k \leq30\}$. This result is shown as a normalized average for each $\sigma$in Fig~\ref{fig6}b. This process is performed to enable comparison of each average result with the results generated by the ViT attention module that will be analyzed subsequently

\begin{figure}[!h]
\centering
    \includegraphics[width=1\linewidth]{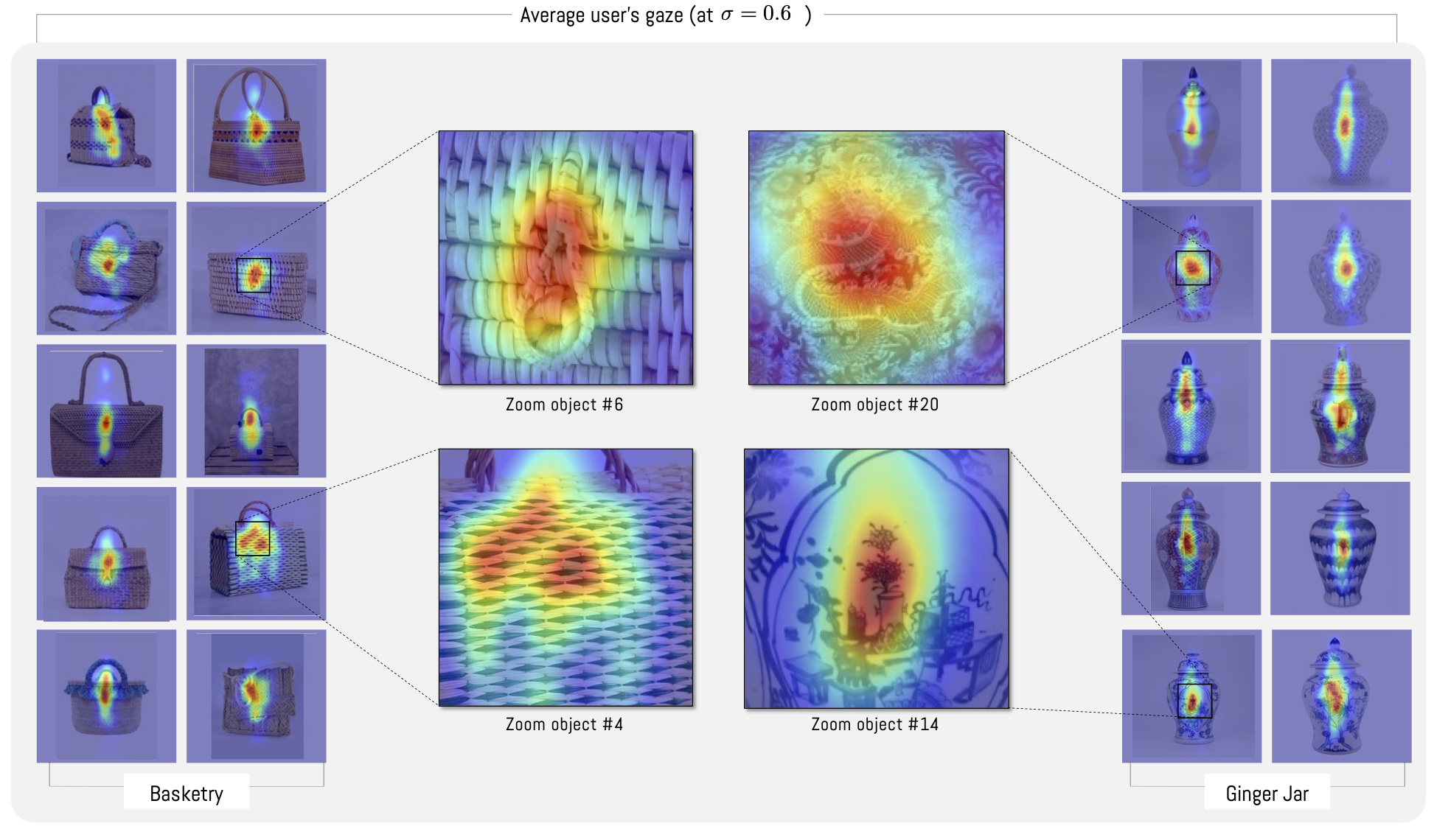}
\caption{Density of users' average gaze for each object for $\sigma=0.6$. Basketry: Zoom object \#4: Detail of region of an object without buckle. Zoom object \#6: Detail of buckle with longer observation time by users. Ginger Jar: Zoom object \#14 vase symbol with the highest amount of observation.}
\label{fig7}
\end{figure}

In general, it is observed that users focus their gaze at the center of objects (see Fig.\ref{fig6} for object \#1). However, when performing this analysis for each object, large differences are observed between objects and their type. For example, in most basketry objects, users focus their gaze longer on the buckle, and do not pay attention to the object's texture or its straps. In the case of jars, a displacement of gaze in a vertical direction is observed (Fig.\ref{fig7}). As in the previous case, the observed data show that there is observation of textures and limited observation of ginger jar objects. This is mainly because observation is related to the task. Since the experiment is free observation, there is no specific task that the user must execute when facing the observed objects. As a result of this process, we observe relevant differences between objects of the same category. For example, if we observe basketry objects \#4 and \#6, we notice that object \#4 does not possess a buckle like the rest of the category (Fig.\ref{fig7}, \#4). In this case, it is observed that the heat map generates a broader observation process since there is no zone or region that causes greater attention over other regions. On the other hand, if we observe the jars, we observe that the average gaze has vertical behavior, which in some cases concentrates on some areas of interest of the object when some emergent pattern exists (Fig.\ref{fig7}, \#14 and \#20).

\subsection*{Images generated by the ViT attention module}
\begin{figure}[!h]
\centering
    \includegraphics[width=1\linewidth]{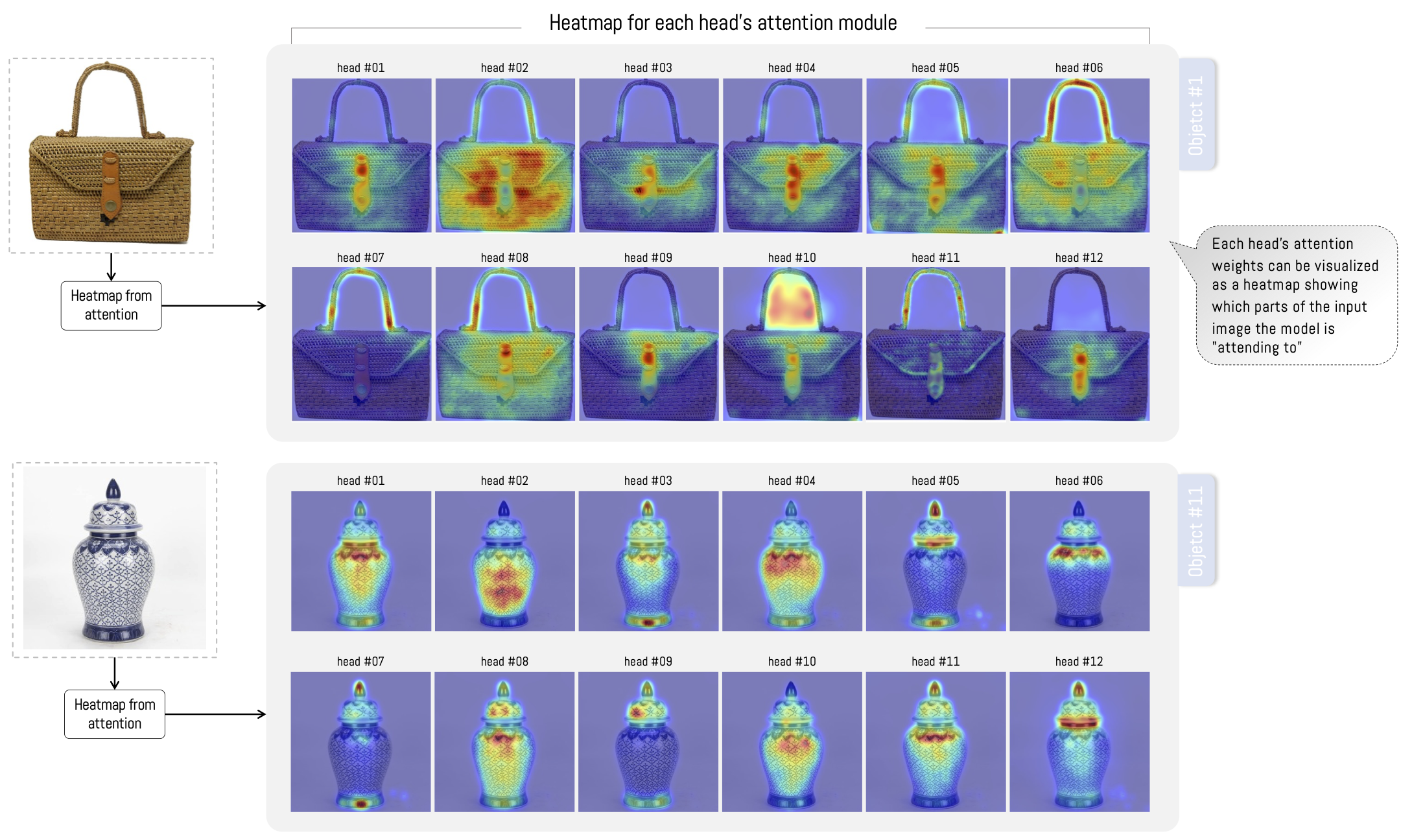}
\caption{12 heatmaps generated by the ViT attention module, both for a basketry-type object and for a jar. Each of the 12 heatmaps represents part of the attention visualization within the algorithm.}
\label{fig8}
\end{figure}

\begin{figure}[!h]
\centering
    \includegraphics[width=1\linewidth]{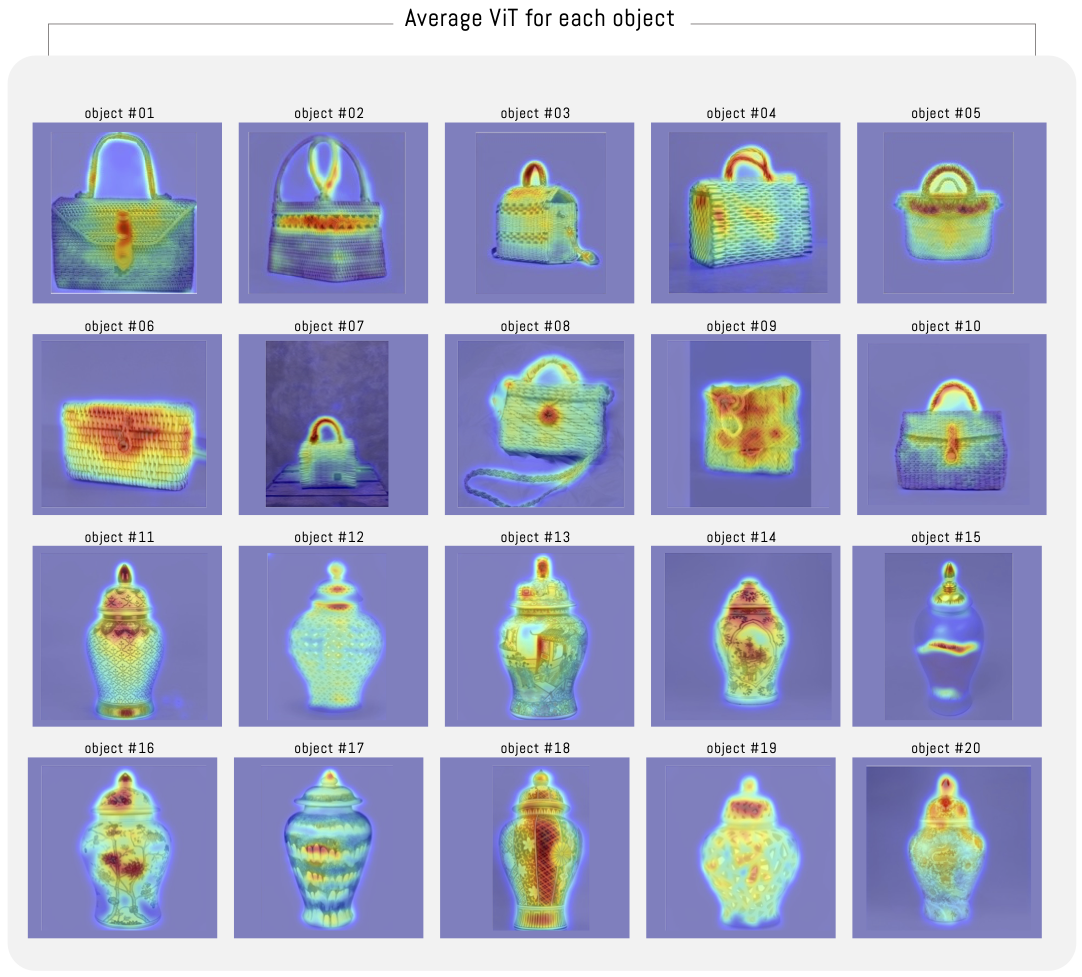}
\caption{Average heatmap of the 12 heads of the ViT attention module for each object in the experiment.}
\label{fig9}
\end{figure}

Fig.~\ref{fig8} presents 12 heatmaps associated with the ViT attention module for objects \#1 and \#11. In the case of basketry-type objects, in some images the ViT attention focuses on texture or on other specific zones of the object such as the buckle or strap. Finally, in some cases a mixture of both is presented (texture and buckle or strap). In the case of ginger jars, each head focuses on a zone of the object, regularly coinciding in the central zone and in some cases on the jar lid. Fig.~\ref{fig9} presents the average of the 12 heatmaps for each object. The average shows how the attention module has more dispersed attention over the entire object. However, in the experiment images we notice relevant differences between the categories used. In this sense, in all basketry objects, the buckle is marked with greater density and then the strap over other zones. In the case of jars, the result is very variable, most of the time the ViT attention module focuses on the lid and in others on some zones with textures and drawings of the jar. In some extreme cases attention is concentrated on some characteristic of the object (see Fig.~\ref{fig9}-object\#15).

\subsection*{Differences between ViT and participant visualization}
\begin{figure}[!h]
\centering
    \includegraphics[width=1\linewidth]{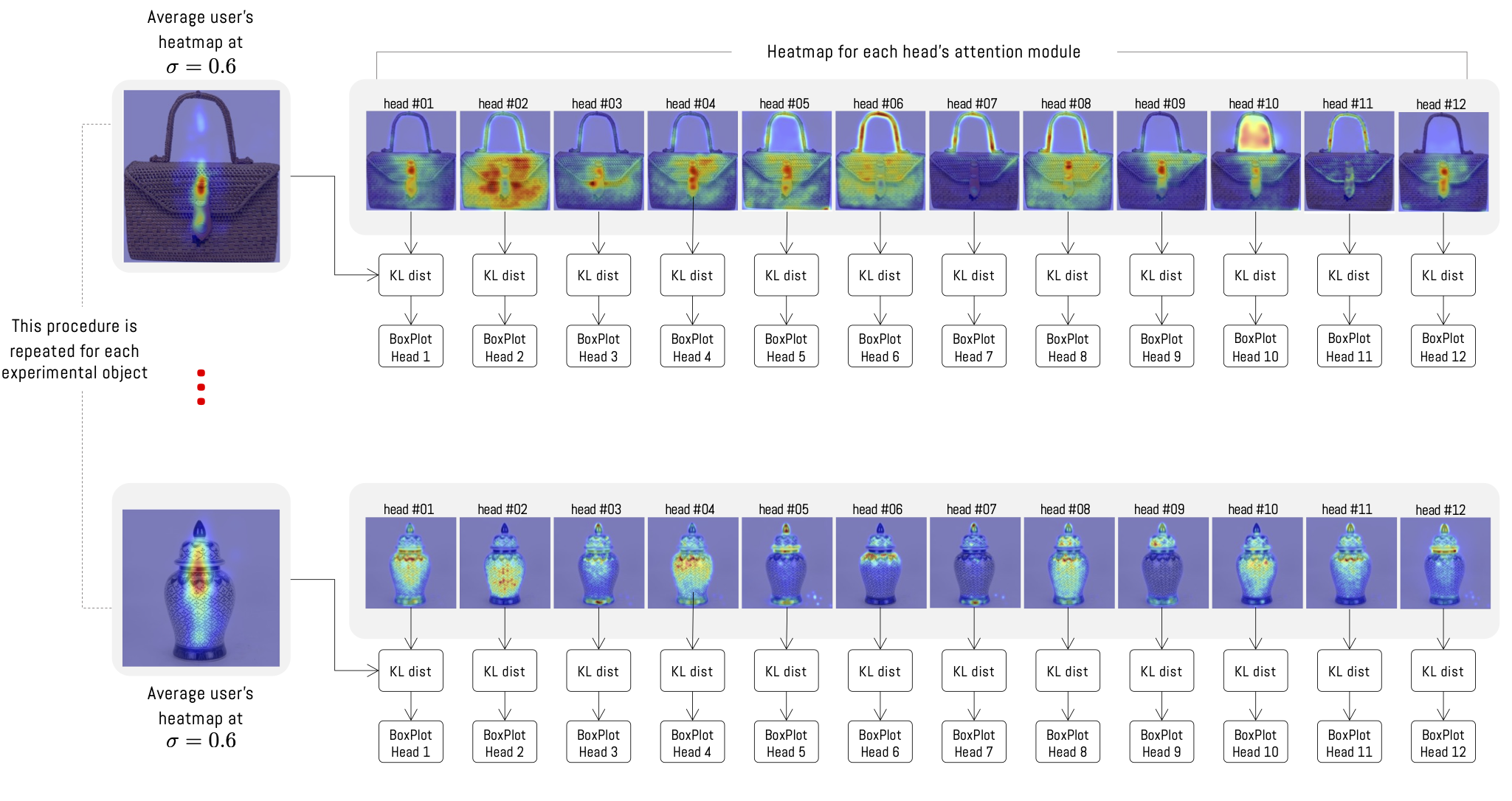}
\caption{KL distance procedure between the 20 objects visualized by people versus each of the 12 heads of the ViT attention module. The figure illustrates the comparison case between object \#4 and object \#15 with respect to the 12 ViT heads.}
\label{fig10}
\end{figure}

To measure the similarity or difference between the heat maps produced by ViT and human attention, we determine the distance using Kullback-Leibler (KL) divergence, which measures the statistical difference between two probability distributions on the same variable \cite{kullback1997information}. To apply this distance, the images to be compared must be normalized (normMixMax [0-1]). This process is performed for each average image visualized by participants (Fig.\ref{fig7}) and each of the ViT heads (12 heatmaps per object) (see Fig.\ref{fig10}). It is important to indicate that we do not determine the distance with respect to the average ViT image shown in Fig.~\ref{fig9}, since we seek to measure if there is any significant difference between any of the ViT heads with respect to the average visual attention of participants. 

\begin{figure}[!h]
\centering
    \includegraphics[width=1\linewidth]{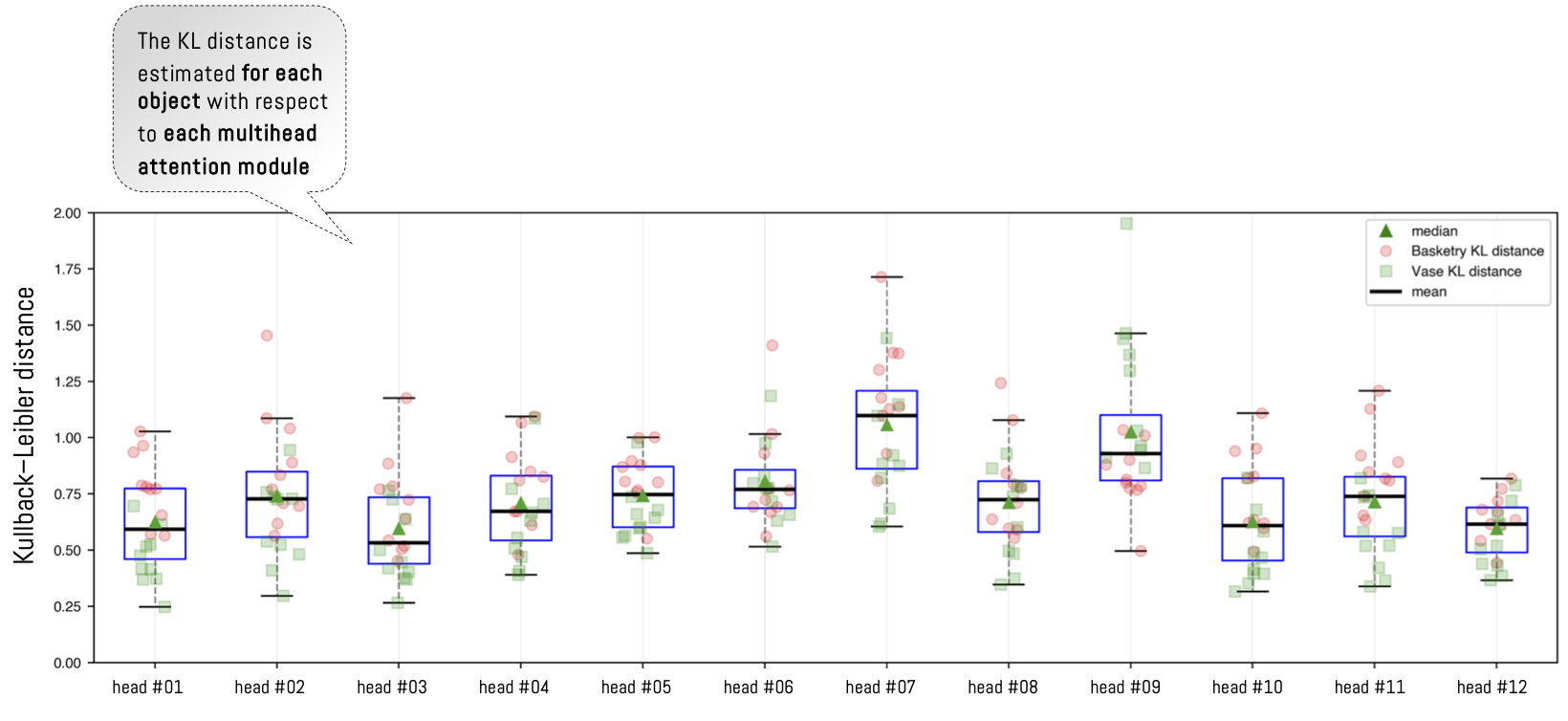}
\caption{Each point in the BoxPlot represents the KL distance between one of the 20 objects and each of the attention heads. In this example, the parameter $\sigma$ is fixed at $0.6$.}
\label{fig11}
\end{figure}

The comparison process is performed for each of the 20 objects visualized by participants (see average visualization in Fig.\ref{fig7}) against each of the 12 attention module heads independently (see example of the process in Fig.\ref{fig10}). In this way, we obtain 12 results for each of the objects (one per head). Since we have 20 objects in the experiment, we total 240 comparisons ($12 \times 20$) given a fixed $\sigma$. In order to visualize this distance for a given $\sigma$, the boxplot in Fig.
\ref{fig11} presents the distance between the average per object versus each head (given $\sigma=0.6$). In general, it is observed that the distances are similar between heads. However, heads \#7 and \#9 consistently have a greater distance than the rest of the heads. Since the experiment considers two categories of objects, it is possible to observe that there are differences from the same head for different categories. This means that heads can have a greater or lesser distance for some classes of objects than for others. For example, head \#1 consistently has a smaller distance for Jar-type objects than for Basketry-type objects. The opposite effect occurs in head \#9 where the head has better performance for basketry-type objects.

\begin{figure}[!h]
\centering
    \includegraphics[width=1\linewidth]{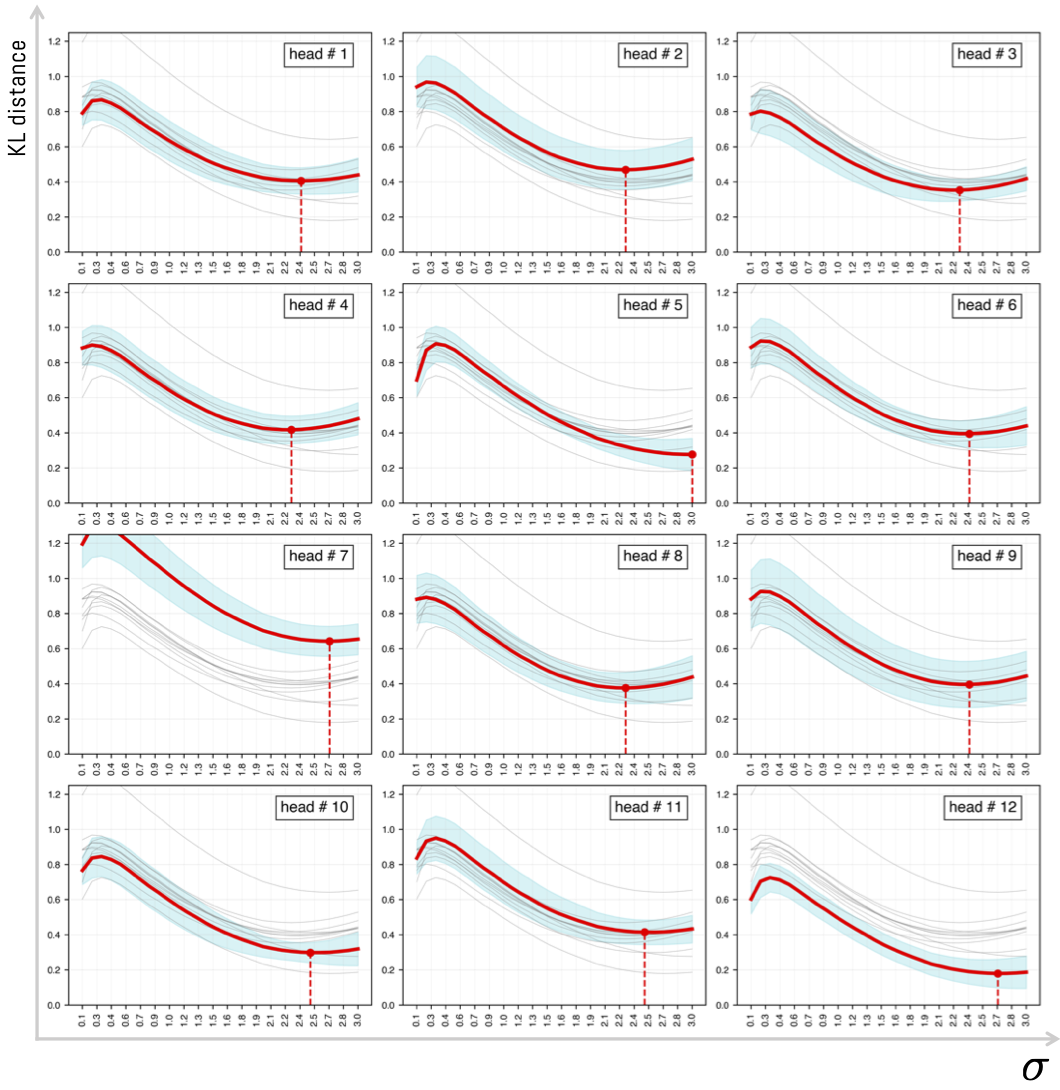}
\caption{Variation of KL distance as the value of $\sigma$ increases for each of the heads. The KL distance has been calculated as the distance between the average visualization of all objects against each ViT head. The blue background corresponds to the sample confidence interval ($95\%$)}
\label{fig12}
\end{figure}

To analyze in detail the effect that variation of the parameter $\sigma$ has, we performed the same experiment varying the values of $\sigma$ between 0.1 and 3.0 with a step of 0.1. In this way, we analyze the average behavior of the KL distance as the parameter increases. Since for each value of $\sigma$, 240 comparisons are necessary, and added to the fact that we generate 30 incremental values of $\sigma$; in total we recorded 7200 combinations ($30\times 240$). The result of this comparison is presented in Fig.~\ref{fig12} separated by head. From the graph it can be seen that although each head may have a different result, in general it is observed that there exists an optimal value whose distance between the head and the average image of participants is minimal. The statistical results indicate that this value is found at $\sigma=2.4\pm 0.03$. In this sense, experimentally we observe that heads \#3, \#10 and \#12 are more similar to the average visualization of people and that, in addition, they possess the lowest variance among the different compared objects. As the parameter $\sigma$ begins to increase, the KL distance increases and at the same time the standard error of the sample (see blue background under each line). It is relevant to note that some of the heads do not have a result close to the average visualization of participants. This is the case of head \#7 and head \#9 where the dispersion is greater with respect to the other heads. This result had already been analyzed previously for the specific case of $\sigma=0.6$
(see Fig.~\ref{fig11}). When analyzing the variation of $\sigma$ over a range of values, the same performance is observed consistently. This result is aligned with previous research where ViT heads do not necessarily fix attention to the same regions that an average human visualization would.

\begin{figure}[!h]
\centering
    \includegraphics[width=1\linewidth]{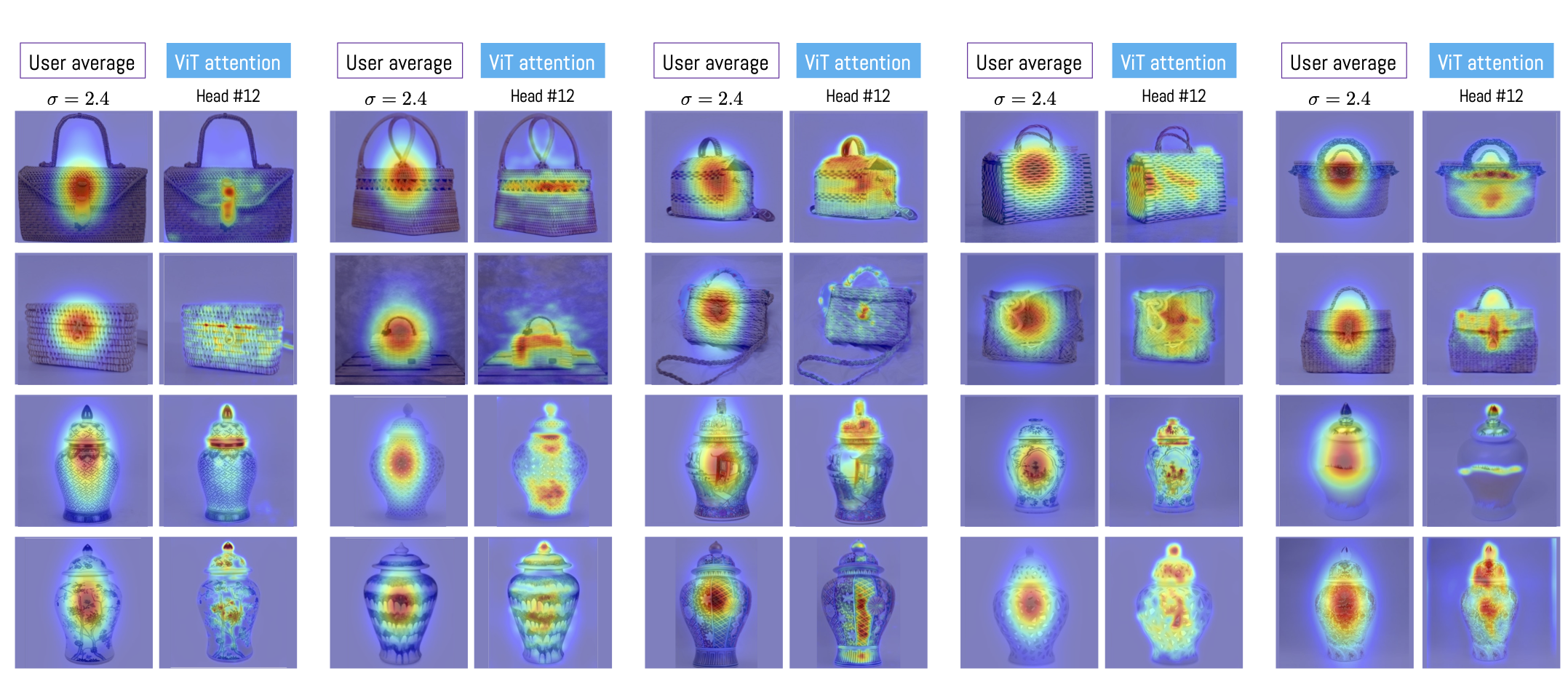}
\caption{Comparison between each average visualization with respect to head \#12 of the ViT attention module. In the case of average visualization, a value of $\sigma=2.4$ has been considered.}
\label{fig13}
\end{figure}

Fig.~\ref{fig13} illustrates the difference between the average visualization of participants and head \#12 of the attention module for each of the 20 objects in the experiment. It is observed how the average visualization of participants tends toward the center of the image, in contrast, ViT head \#12 detects specific zones of the object, which in some cases coincides with the average visualization of participants. This example illustrates how one of the 12 heads tends to approximate the average visualization of experiment participants

\begin{figure}[!h]
\centering
    \includegraphics[width=1\linewidth]{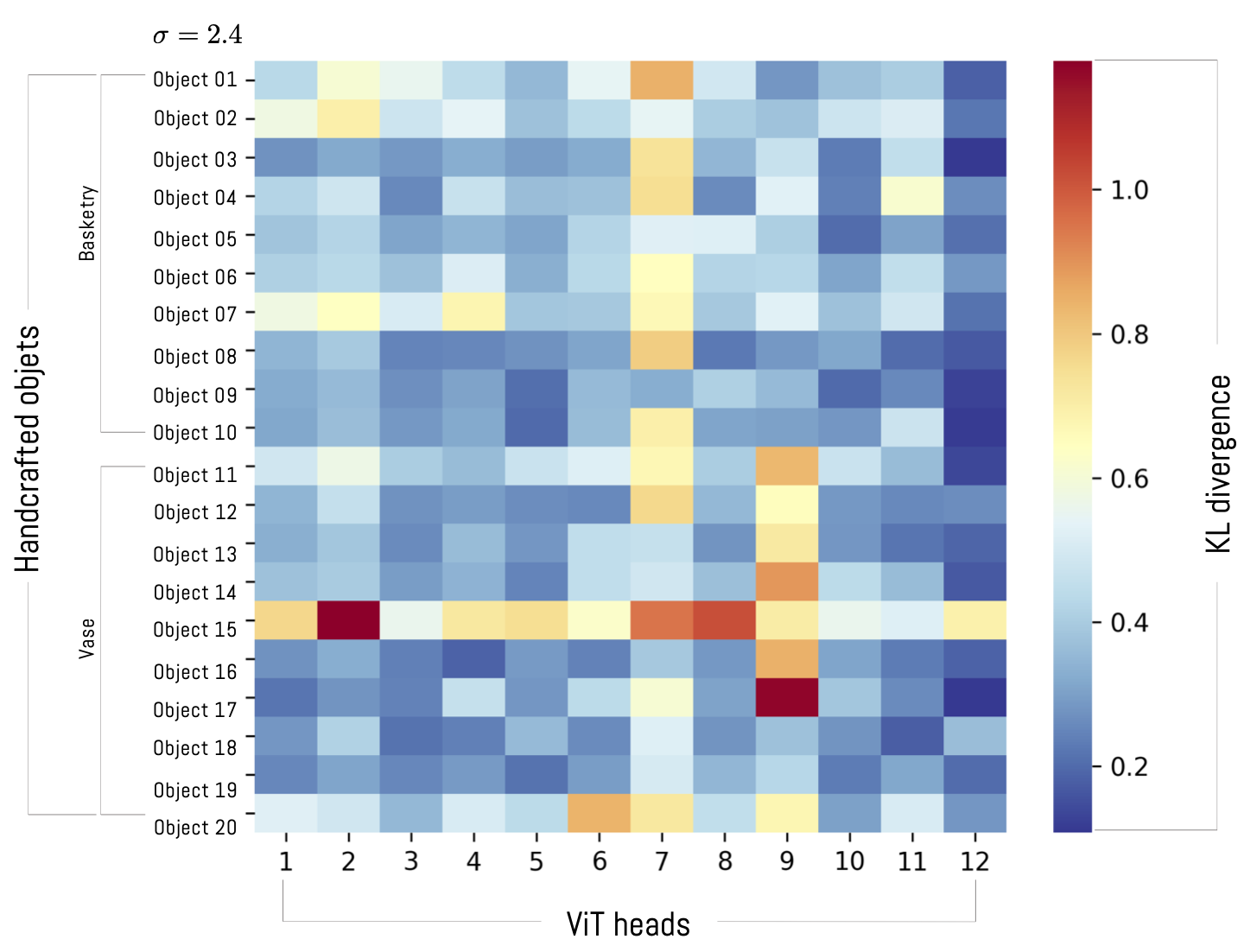}
\caption{Heatmap with the KL distance between each object and each of the ViT attention module heads. This result has been calculated with $\sigma=2.4$}
\label{fig14}
\end{figure}

The KL distance with respect to each object and each attention module head are presented in Fig.~\ref{fig14}. In this graph it is observed how head \#12 presents a difference from the other heads since it consistently has the lowest KL distance. However, it is interesting to note that the distance is lower for basketry objects versus ginger jars. On the contrary, heads \#7 and \#9 have the greatest distance with respect to the analyzed objects. In particular, head \#9 tends to have worse performance for jars than for basketry. This result highlights the difference and focus that the attention module possesses, and reveals that analyzing each head independently allows understanding the adjustment process of each head in relation to the average value of the ViT.

\begin{figure}[!h]
\centering
    \includegraphics[width=1\linewidth]{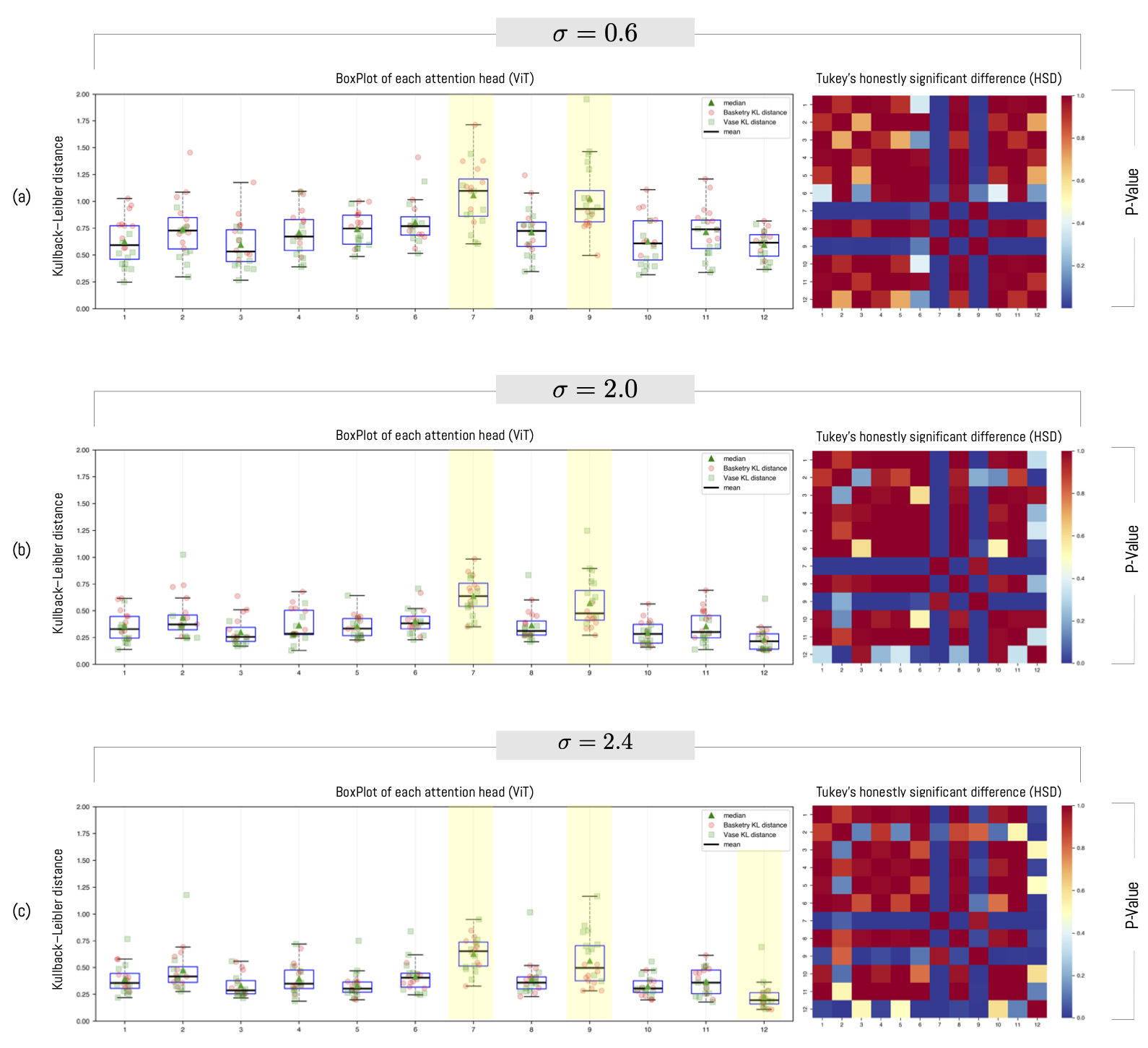}
\caption{Difference between Tukey Honestly Significant Difference (HSD) to measure the difference in means between attention module heads with three variants of $\sigma$.}
\label{fig15}
\end{figure}

To analyze the distance relationship between the different heads, we employed the Tukey Honestly Significant Difference (HSD) test after checking with an ANOVA test that there is significant difference between the means (see Fig.\ref{fig15}). For this we fixed the value of $\sigma=2.4$ and applied the test between all heads. The results indicate that heads \#7, \#9 and \#12 are statistically different from the rest of the heads, given that their p-values in most cases is less than 5\%. These results, together with those presented in Fig.~\ref{fig14}, allow us to affirm that heads \#7 and \#9 are those with the greatest distance to human visualization and head \#12 with the smallest distance.

\section*{Discussion}
Computer vision has achieved great advances in the development of attention mechanisms based on human vision \cite{guo2022}. Thus, vision transformers have demonstrated high potential in this field for different tasks, equaling or even surpassing Convolutional Neural Networks (CNN), which present limitations compared to human attention\cite{baker2020,caron2021,wloka2019}. Our research continues to deepen the relationship between human attention and that performed by unsupervised ViTs\cite{mehrani2023,yamamoto2025a}. For this, hypotheses were established that there are no significant differences between the attention produced by a ViT model and human attention (H1), so it can be an applicable technology in the design and creation process of artisanal products for detecting elements with greater aesthetic attractiveness (H2). To respond to the hypotheses, objectives were established to statistically determine the correlations between both attention mechanisms and analyze visual interest points in artisanal objects through these attentions.

The methodology established to achieve the objectives was organized in three stages (data preparation, modeling and evaluation), where a dataset of images of two typologies of artisanal objects was constructed, 10 bags made through basketry and 10 ginger jars, with different stylistic characteristics where, in the case of bags made with basketry, for their selection structural unity was sought, with polygonal forms predominating, with small curvilinear contrasts in handles or closure clasps. The bags possess similar textures and colors, including some differentiating examples with small details of different colors and materials to find variations in attention between images. In the case of ginger jars, curvilinear forms and verticality predominate. These crafts possess different aesthetic elements in their decoration, such as floral motifs or naturalistic representations, as well as textures caused by decorative or structural frameworks.  

These artisanal objects were visually evaluated by 30 people (9 women and 21 men), with an average age of 24.6 years (SD=3.97), using an Eye Tracker that allowed recording the participants' observation. With the obtained data, 600 heatmaps were generated, which were normalized and averaged using a two-dimensional Gaussian distribution and altering the $\sigma$ parameter, allowing their subsequent comparison with the results obtained from the ViT.

The results obtained from the participants' viewing in the experiment coincide with what was expected following the stylistic characteristics of the objects \cite{schoenmaekers2025,stapel2021}. In the case of bags that have closure buckles, they concentrate attention, while those objects that do not have this element, attention disperses across surfaces, especially on textures different from the general structure. This occurs because there are differentiating forms from the set that help visual attention. In this case, the circular forms of the closures create a contrast with the general rectilinear forms of the bag structures. In objects viewed without this element, textures different from the general set possess different characteristics of framework and colors that cause greater visual attention. On the contrary, ginger jars provoke ascending and descending attention, conducted by the morphology of these products. As with basketry craft objects, central attention predominates, which coincides with previous studies on aesthetic preferences and spatial composition~\cite{sammartino2012}, although deviated attentions are detected when the object possesses some decorative element, especially the images of ginger jars that have floral motifs and represented people (Fig.\ref{fig7}), which is consistent with previous works on Object-based visual attention \cite{cavanagh2023,sun2003}.

Most of the images used in the experiment do not have a background, since it is a context that can cause distractions by having semantic content \cite{luo2020}. Only one image of a basketry bag with background and another element (a table where the object rests) has been preserved to analyze if these elements cause distractions in attention. The results do not show significant attention from humans to these elements.

The same image dataset was analyzed using Vision Transformer (ViT), with pre-training with DINO (Self-DIstillation with NO Labels). This architecture generates 12 independent distributions (heads). This implies the creation of 240 heat maps (12 for each craft object). As observed in Fig.\ref{fig9}, the attention performed by ViT is deployed over the entire surface of the object, coinciding with previous research that indicates the tendency toward globalized attention in images \cite{yamamoto2025a}. This is consistent with the research by \cite{mehrani2023}, who indicate that ViTs do not perform selective attention, but rather groupings by element characteristics, acting as a horizontal relaxing labeling and bottom-up processing in human attention. However, unlike our work, the researchers do not collect and use data on human attention. This allows us to observe certain similarities, since both ViT and human observation, on basketry bags, focus on buckles when this element is present. However, differences are observed in attention to jars, where ViT does not follow a verticality as in the attention of experiment participants.

To better understand the results, the attention produced by each head of the ViT used was analyzed, following previous research such as \cite{li2023} or \cite{michel2019}, seeking to better understand the internal functioning of the model. For this, normalization (normMixMax) was performed on each average image of participants and on each ViT head, and subsequently the parameter $\sigma$ was altered. Experimentally, the Kullback-Leibler (KL) distance comparison was established with $\sigma\in\{0.1,0.2,0.3,...,3.0\} = \{0.1\times k | \space k \in \mathbb{N},\space 1\leq k \leq30\}$ (see Fig.~\ref{fig6}).

The results show that there are heads more correlated with human attention than others. Thus, it is observed that, in the average of all attentions, heads \#3, \#10 and \#12, are the most consistent according to KL distance. The greatest dispersion is found in heads \#7 and \#9. This result is partially consistent with the study \cite{yamamoto2025b}, where they indicated that the attentions most similar to human ones (in adults) occurred in intermediate heads, and shows that not all heads are relevant for the task \cite{li2023}.

To understand in detail the effect of variation in the parameter $\sigma$ (from 0.1 to 3.0), 7200 analyses were performed corresponding to the 30 variations of 0.1 by the 240 KL distances. The results show that statistically, the optimal value is found at $\sigma=2.4\pm 0.03$, with head \#12 being the closest to the average attention of participants, being the most consistent in all analyses (Fig.~ \ref{fig14}). Similarly, heads \#7 and \#9 have the greatest dispersion. These results are consistent with the previous experiment with $\sigma=2.4$. To verify these results statistically, the Tukey Honestly Significant Difference (HSD) test was used, and it was concluded that heads \#7, \#9 and \#12, are significantly different from the rest, with $p \leq 0.05$.

\section*{Limitations and future directions}
Among the main limitations of this research, the reduced number of participants in the experiment stands out, so the sample cannot be considered representative and makes it difficult to perform experiments taking into account sociodemographic data and gender perspective. Therefore, our intention is to continue increasing the sample in the future. Additionally, we intend to expand the experiment to other geographical and cultural contexts, since, like aesthetic experience, the cultural context of observers is determining in visual attention along with other factors, such as age, gender, educational level or field of study \cite{boduroglu2009,chamberlain2015,suh2023}. This is why, in this research, such information has begun to be compiled, but has not been included in the methodology due to its limited relevance (e.g., observe that, in the participants of this research, the number of men is significantly higher than women to make a comparison). Undoubtedly, in the future, these demographic data and the incorporation of gender perspective will allow greater thoroughness in analyses and reach more significant conclusions, where we can hypothesize if any gender is closer to attention performed with ViT, or if there are differences between observers with different levels of education or areas of knowledge, making a comparison equally with ViT.

Despite the limitation of the number of participants, numerous data have been extracted through the applied methodology. Let us remember that each participant viewed 20 objects, which means, among other possibilities, obtaining 600 attention heatmaps. This also increases with the analysis of each object by model head, reaching 240 ViT attention heatmaps. These data increase significantly when analyzing the variation of $\sigma$, which required 7200 combinations to reach results on the optimal value in this parameter.

The results of this research reveal greater correlation between certain ViT heads and human attention, but this finding requires deeper analysis to understand the model's mechanisms. For example, the importance of each head could be studied following Michel et al.~\cite{michel2019}, who found that disabling certain heads did not significantly affect performance. Using this methodology, we could determine whether the heads that showed significance in our experiment are actually fundamental to the attention model.

Another future line of action involves analyzing the correlation between purchase intention and visual attention patterns for the same product. This experimental approach would require observers to indicate their purchase intention while viewing objects, with simultaneous recording of response times and eye fixations \cite{gidlof2017,jiang2016,li2024,liu2020}. Such an approach would enable us to determine more precisely whether aesthetic interest serves as a determining factor in purchase decisions \cite{isham2013}.

\section*{Conclusiones}
This study has allowed for deeper exploration of the relationship between human visual attention and that generated by unsupervised Vision Transformer (ViT) models, applied to aesthetic analysis of artisanal objects. The obtained results demonstrate that, while there are differences in the spatial distribution of attention between both mechanisms, relevant coincidences are identified in certain visual elements, such as buckles in basketry bags, suggesting potential use of ViT in product design processes to predict zones of interest.

The detailed analysis of ViT heads has revealed that some of them (specifically head \#12) present greater correlation with human attention, especially when the parameter $\sigma$ is adjusted to the analyzed optimal value $(\sigma=2.4\pm0.03)$. These findings reinforce the hypothesis that certain ViT components can approximate human attentional behavior, opening new possibilities for their application in creative production contexts. Despite methodological limitations, such as the reduced sample size and lack of sociodemographic analyses, the research has generated a solid foundation for future investigations. Future research lines include studying the functional relevance of each ViT head, as well as incorporating sociodemographic variables that allow for more exhaustive understanding of individual differences in visual attention.

Overall, the obtained results partially support the viability of using ViT models as complementary tools in aesthetic and perceptual analysis of artisanal products, contributing to the dialogue between artificial intelligence and human creativity.

\section*{Data and code}
Data and code is available at \url{https://github.com/mlacarrasco/human_versus_vit}

\section*{Acknowledgments}
We would like to express our sincere gratitude to the Neuroscience Laboratory of the School of Psychology at Universidad Adolfo Ibáñez (UAI) for providing the facilities, equipment, and technical support that made this research possible.. We also extend our appreciation to all the participants who voluntarily contributed their time to this study. This research was conducted with the approval of the ethics committee of Universidad Adolfo Ibáñez (certificate 57/2023).


%
%
%

\bibliographystyle{plos2015}  
\bibliography{references}     

\end{document}